\pdfoutput=1
\documentclass[letterpaper, 10 pt, conference]{ieeeconf}

\IEEEoverridecommandlockouts
\usepackage{cite}
\usepackage{amsmath,amssymb,amsfonts}
\usepackage{algorithm}
\usepackage{algpseudocode}
\usepackage{graphicx}
\usepackage{booktabs}
\usepackage{array}
\usepackage{multirow}
\usepackage{tabularx}
\usepackage{textcomp}
\usepackage{url}
\usepackage{xcolor}
\usepackage[normalem]{ulem}
\usepackage{kotex}

\usepackage{xspace}
\usepackage[hidelinks,bookmarks=false]{hyperref}

\newcommand{\methodabbr}{IDP\xspace}
\newcommand{\rev}[1]{#1}
\newcommand{\termfix}[1]{#1}
\makeatletter
\let\reviewruledstyle\fs@ruled
\renewcommand{\fs@ruled}{\reviewruledstyle
  \def\@fs@pre{{\color{black}\hrule height.8pt depth0pt}\kern2pt}
  \def\@fs@mid{\kern2pt{\color{black}\hrule}\kern2pt}
  \def\@fs@post{\kern2pt{\color{black}\hrule}\relax}
  \def\@fs@capt##1##2{{\color{black}{\bfseries ##1} ##2\par}}}
\makeatother

\graphicspath{{figures/}}

\title{\LARGE \bf Interactive Diffusion Policy: Safe Reactive Avoidance through
       Proximity Sensing in Human-Robot Shared Environments}

\author{Yeong Gwang Son$^{*}$, Hyunchang Kang$^{*}$, Youngbin Song, Hyunjoon Cho, and Hyouk Ryeol Choi$^{\dagger}$%
\thanks{$^{*}$These authors contributed equally to this work.
$^{\dagger}$Corresponding author.}%
\thanks{The authors are with the Department of Mechanical Engineering,
Sungkyunkwan University, Republic of Korea.}}

\begin{document}

\maketitle
\thispagestyle{empty}
\pagestyle{empty}

\begingroup
\color{black}
\begin{abstract}
Manipulators sharing a workspace with humans must avoid approaching obstacles while completing their tasks. Such avoidance requires obstacle detection, but camera-based detection can be hindered by occlusion. Avoidance can also move the robot outside the demonstration distribution, making task resumption difficult for a learned policy. These challenges call for a method that integrates reactive avoidance with recovery from out-of-distribution states.

In this paper, we propose Interactive Diffusion Policy (IDP), which connects task execution, reactive avoidance, and recovery. Diffusion Policy generates a reference target, and a lower level RMPflow controller integrates target tracking and obstacle avoidance using distance information from proximity sensors. Robot mounted capacitive proximity sensors provide local distance information to compensate for camera observation limitations. Sensor experiments confirmed that the proximity sensors supported avoidance even for occluded and transparent objects that a depth camera failed to detect. To resume the task after avoidance, recovery data are collected by human guidance from off-path states caused by actual RMPflow avoidance. These data are used to train the dynamics model of the Latent Policy Barrier (LPB), which guides task resumption after avoidance. Manipulator experiments under various obstacle conditions showed that the robot resumed the task after avoidance and achieved a task success rate of 91.3\%. These results show that IDP integrates collision avoidance and task resumption, enabling task execution that considers both safety and task capability in human shared workspaces. \textnormal{\urlstyle{same}\url{https://interactive-diffusion-policy.github.io/}}
\end{abstract}
\endgroup

\begingroup
\color{black}
\section{Introduction}
In manufacturing, logistics, and service environments, robots
increasingly share workspaces with humans rather than operating in
physically separated areas. In these environments, humans may
unexpectedly enter a robot workspace or cross its motion path. Robots
must therefore respond to changes in their surroundings in real time
to prevent collisions. However, repeatedly slowing down or
interrupting a task for safety reduces task continuity and efficiency.
Robots operating in dynamic human shared environments must therefore
respond safely to human approaches while continuing and completing
their assigned tasks.

Imitation learning is effective at learning complex robot manipulation
from demonstrations, and visuomotor policies such as Diffusion Policy (DP)
can generate actions for a variety of manipulation
tasks~\cite{chi2023diffusion}. However, these policies learn task
behaviors from the given demonstrations and do not inherently guarantee
safe responses to unexpected human approaches or environmental changes
during execution. Accordingly, previous studies have introduced methods
that evaluate motions generated by learned policies and restrict or
replace motions that may lead to collisions~\cite{pacs2024,rail2024}.

However, they focus on preventing unsafe execution and do not jointly
address the detection of an approaching person or object under visual
occlusion, spatial avoidance, and subsequent task resumption. In
particular, reactive avoidance can move the robot away from the
demonstrated trajectory and into an out-of-distribution (OOD) state that was
not observed in the demonstrations. From this state, the learned policy
may fail to continue the original task. Therefore, dynamic workspaces
shared by humans and robots require a method that connects reliable
detection of occluded approaches with reactive avoidance and task
recovery after avoidance.

\begin{figure}[!t]
    \centering
    \includegraphics[width=\columnwidth]{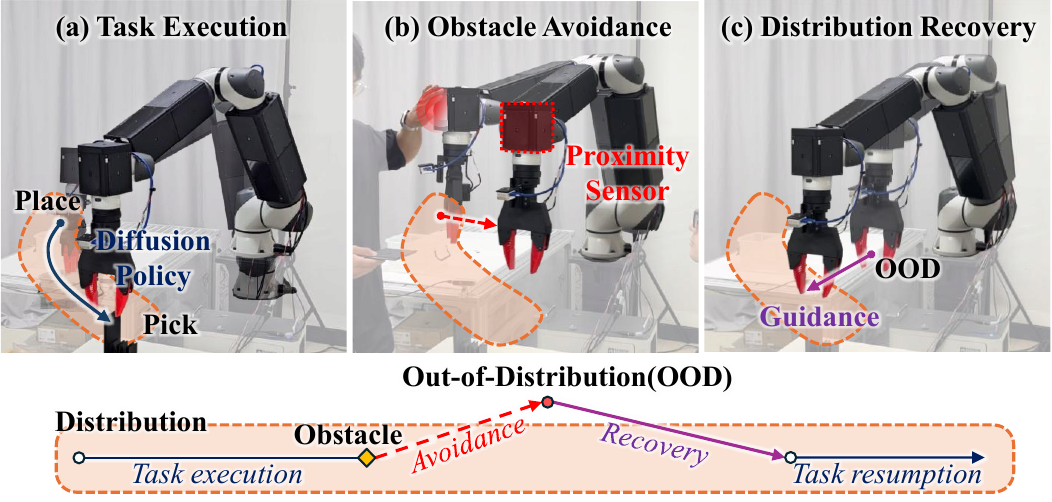}
    \caption{Overview of nominal policy execution, avoidance by
        RMPflow, and distribution recovery guided by LPB.}
    \label{fig:teaser}
\end{figure}

This study proposes Interactive Diffusion Policy (IDP), illustrated in
Fig.~\ref{fig:teaser}, to address these problems. IDP separates task
observation from safety perception by using vision for Diffusion Policy
observations and distance measurements from robot-mounted capacitive
proximity sensors for reactive safety
control~\cite{yim2024emtof,kang2025adaptive,moon2021dual}. Through
proximity field sensing, these sensors detect nearby people and objects
from changes in the electric field around the robot surface, making
them suitable for close human--robot interaction. To use these distance
measurements for avoidance during task execution, IDP employs
RMPflow~\cite{cheng2020rmpflow} as a low-level controller that
continuously combines tracking of reference targets generated by
Diffusion Policy with obstacle avoidance based on proximity measurements.

To support task resumption from states outside the learned task
distribution after avoidance, this study builds on the dynamics model
of Latent Policy Barrier (LPB)~\cite{lee2025latentpolicybarrier}.
While the original LPB constructs out-of-distribution data from rollouts
of partially trained policies, this study augments the dynamics model's
training data with rollouts that contain actual proximity-based
RMPflow avoidance and the subsequent resumption, in which a human
operator guides the robot when the policy cannot resume on its own. The model thus learns state transitions following avoidance
and guides Diffusion Policy from the resulting off-path states back
toward the task distribution.

The main contributions of this work are as follows.
\begin{enumerate}
    \item \textbf{Safety-oriented imitation learning framework.}
    IDP combines task execution guided by Diffusion Policy with reactive
    avoidance by RMPflow using proximity data.

    \item \textbf{Proximity-based safety perception.}
    Proximity sensors detect approaching objects and provide information
    for avoidance when the camera view is occluded.

    \item \textbf{Task recovery after avoidance.}
    The LPB dynamics model is trained with recovery data from actual
    avoidance to improve task resumption.

    \item \textbf{Real-robot experiments.}
    Experiments with a robot manipulator evaluate collision avoidance,
    task recovery, task completion, and the effect of recovery data.
\end{enumerate}
\endgroup
\section{Related Work}
\subsection{Proximity Sensing for Robots}
\begingroup
\color{black}
Capacitive proximity sensors detect nearby people and objects by sensing
changes in the surrounding electric field. This proximity field sensing
has been fused with time-of-flight (ToF)~\cite{yim2024emtof} and
LiDAR~\cite{kang2025adaptive} to support safer human--robot interaction.
Moon et al.~\cite{moon2021dual} combined capacitive and inductive sensing
for manipulator proximity detection and self-sensing compensation.
This study uses proximity sensing to complement visual perception with
local distance measurements of approaching people and objects,
supporting reactive avoidance during task execution.
\endgroup
\subsection{Imitation Learning from Human Demonstrations}
\begingroup
\color{black}
Imitation learning has shown that a robot can acquire the intuition a
person brings to manipulation and carry out a wide range of tasks. The
policy is trained on data that record how a person performs the task,
and recent visuomotor policies map raw observations directly to
actions, either by regressing action chunks~\cite{act2023} or by
generating them through iterative denoising~\cite{chi2023diffusion}.
Later variants extended these policies with force and compliance
feedback~\cite{compliantdp2024, acp2024, umift2026}, widening the
range of tasks a learned policy can carry out. These studies, however,
are evaluated on task success and do not address physical safety,
since the workspace is assumed to contain only the objects represented
in the demonstrations.
\endgroup

\subsection{Sensor-Based Safety Control}
\begingroup
\color{black}
Proximity measurements and visual obstacle information have been used
in control and motion planning for manipulator collision avoidance.
Moon et al.~\cite{moon2021dual} integrated proximity sensors with a
manipulator to enable real-time obstacle avoidance for safe
human--robot interaction. cuRobo~\cite{sundaralingam2023curobo}
generates collision-free trajectories through GPU-parallel optimization
using obstacle representations that include distance fields constructed
from depth images. RMPflow~\cite{cheng2020rmpflow} geometrically
combines motion policies for target tracking and obstacle avoidance,
while DRP~\cite{drp2025} uses point-cloud-based DCP-RMP to adjust the
joint-space goals supplied to a learned motion policy in response to
moving obstacles.

Other studies impose safety conditions during execution or trajectory
generation to improve the safety of learned policies and generative
models. PACS~\cite{pacs2024} and RAIL~\cite{rail2024} use reachability
analysis to verify safety. PACS brakes along the policy's intended
path, whereas RAIL switches to a verified stopping plan when safety
verification fails. CAPE~\cite{cape2025} uses the unexecuted segment
of the previous plan as a prior and guides denoising with a collision
cost computed from obstacle point clouds.
SafeDiffuser~\cite{xiao2023safediffuser} and
SafeFlow~\cite{safeflow2025} apply barrier-function-based safety
constraints during diffusion and flow-matching generation,
respectively. This study combines reference targets generated by a learned
policy with distance feedback from robot-mounted proximity sensors in
a low-level RMPflow controller, enabling target tracking and obstacle
avoidance during task execution.
\endgroup

\subsection{Handling Out-of-Distribution States in Imitation Learning}
\begingroup
\color{black}
A behavior-cloned policy acts on states induced by its own previous
actions, so small imitation errors compound and drive the robot into
states absent from the demonstrations~\cite{dagger2011}. Existing
remedies widen the training distribution by aggregating expert labels
on the states the learner visits~\cite{dagger2011}, by injecting noise
while the expert demonstrates~\cite{dart2017}, or by collecting human
interventions during deployment~\cite{sirius2023, crdagger2025}, while
runtime monitors flag the states a policy can no longer
handle~\cite{sentinel2024}. The Latent Policy Barrier
(LPB)~\cite{lee2025latentpolicybarrier} instead recovers during
execution, treating the latent embeddings of the expert demonstrations
as an implicit barrier and steering the denoising process with a
learned latent dynamics model whenever the current embedding falls
outside it.

In all of these methods the out-of-distribution state originates from
the policy's own error. We adopt LPB, but train its dynamics model on
states produced by actual RMPflow avoidance.
\endgroup

\begingroup
\color{black}
\section{Proximity Sensor}
\label{sec:proximity}
Proximity sensor modules are mounted on the RB10 links, as shown in
Fig.~\ref{fig:sensor}. The modules follow the capacitive and ToF design
of Yim et al.~\cite{yim2024emtof} and are adapted for mounting on the
RB10. The resulting distance estimates are supplied to the obstacle
term in RMPflow.

\subsection{Principle of Proximity Sensor}
Capacitive sensing detects nearby objects through changes in the
electric field around a sensing electrode. When a human body part
approaches the sensor surface, the effective capacitance between the
electrode and the target changes. The simplified model used by
Yim et al.~\cite{yim2024emtof} relates capacitance to distance as
\begin{equation}
    C = \epsilon_0 \epsilon_r
        \frac{2 A_1 A_2}{d(A_1 + A_2)}
    \label{eq:cap_distance}
\end{equation}
Here, $C$ is the measured capacitance, $A_1$ and $A_2$ are the areas
of the sensing electrode and target, and $d$ is the distance between
them. The symbols $\epsilon_0$ and $\epsilon_r$ denote the permittivity
of free space and the relative permittivity, respectively. According
to~\eqref{eq:cap_distance}, capacitance increases as the target
approaches the sensor.

In practice, the capacitance response depends on the target's size,
shape, material, and grounding. Therefore, raw capacitance cannot be
used directly as a distance measurement. In the region where the
capacitive and ToF sensing ranges overlap, paired measurements are used
to fit the empirical relation $C = a e^{bd}$. Inverting this relation
gives the calibrated distance as
\begin{equation}
    d = \frac{1}{b}\left(\ln C - \ln a\right)
    \label{eq:self_curve_fitting}
\end{equation}
Here, $a$ and $b$ are fitting coefficients estimated from paired ToF
distance and capacitance measurements. ToF measurements provide the
reference distance for calibration. Details of the sensor design and
fitting procedure are given in~\cite{yim2024emtof}. Each sensor module
provides a local distance estimate around the robot surface.

\begin{figure}[t]
    \centering
    \includegraphics[width=\columnwidth]{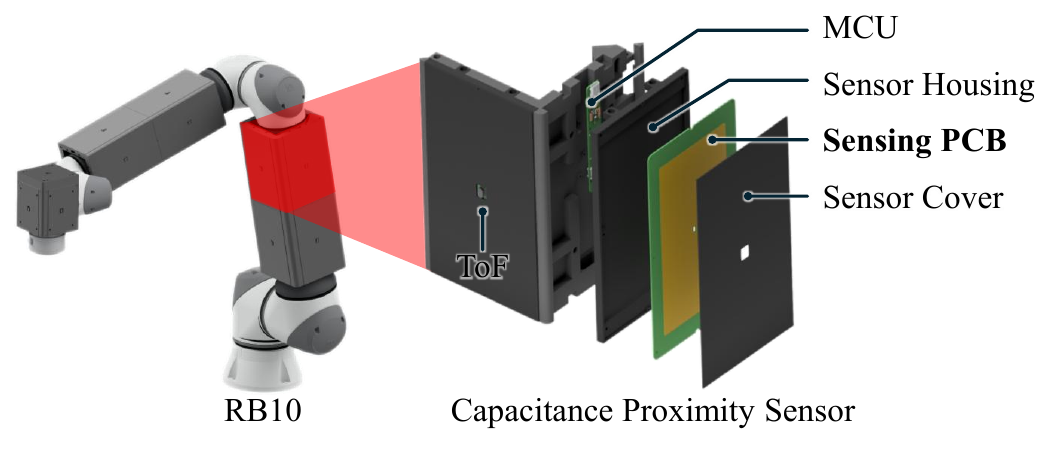}
    \caption{Proximity sensor placement on the RB10 and an enlarged
        view of the highlighted link (left), with an exploded view of
        the sensor module showing the ToF sensor and capacitive
        sensing PCB (right).}
    \label{fig:sensor}
\end{figure}

\subsection{Sensing Performance}
\label{sec:sensor_perf}
The distance estimation performance of the proximity sensor was evaluated
using the moving jig shown in Fig.~\ref{fig:sensor_perf}(a). A metal
specimen mounted on the jig was moved toward and away from the sensor.
The reference distance and the distance estimated by the sensor were
recorded throughout. In the upper plot of Fig.~\ref{fig:sensor_perf}(b),
the black dashed line shows the reference distance, and the red solid
line shows the sensor estimate. The blue solid line in the lower plot
shows the error between them. The sensor estimate followed the repeated
changes in reference distance, with an error of approximately $\pm20$~mm.

Fig.~\ref{fig:sensor_compare} compares the obstacle information
provided to the avoidance controller by a depth camera and by proximity
sensors mounted on the robot. The same RMPflow controller was used,
and only the sensing input was changed. In the suddenly appearing
obstacle condition with an unobstructed camera view in
Fig.~\ref{fig:sensor_compare}(a), both approaches avoided the obstacle.
In Fig.~\ref{fig:sensor_compare}(b), the robot blocked the camera's view
of the approaching object, causing a collision under camera control.
In contrast, the proximity sensors detected the object within the
sensing field around the robot links, allowing the controller to avoid
it. In Fig.~\ref{fig:sensor_compare}(c), the transparent glass was not
detected in the depth image, and camera control resulted in a collision.
The controller using proximity sensor input detected and avoided the
glass.

These results show that proximity sensors provide distance information
about objects approaching the robot and support avoidance when the camera
view is blocked or an object is not detected in the depth image.

\begin{figure}[t]
    \centering
    \includegraphics[width=\columnwidth]{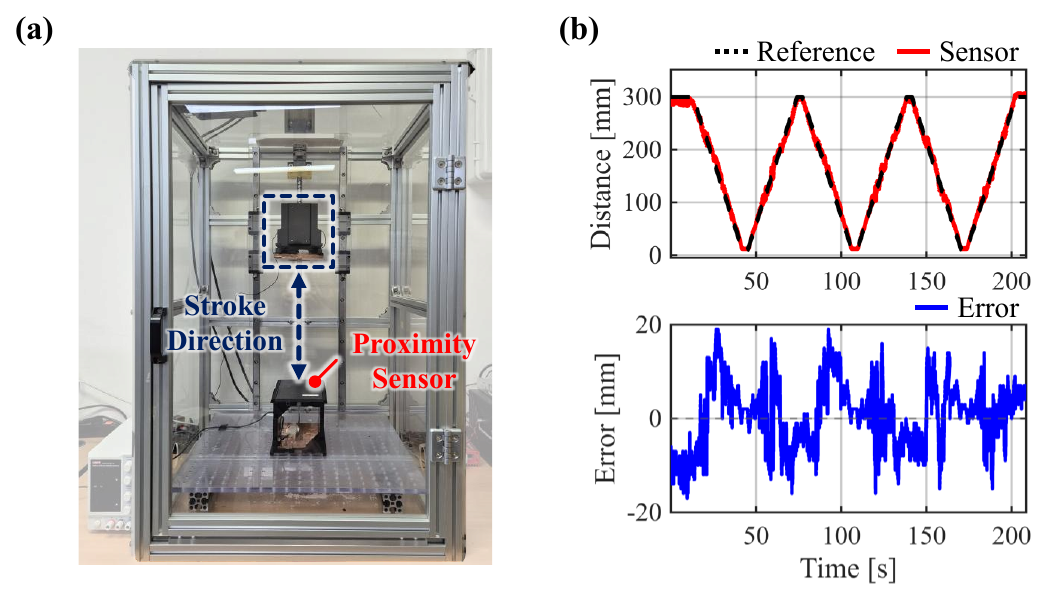}
    \caption{Distance estimation of a metal specimen with a proximity sensor.
        (a) Test jig moving the specimen toward and away from the sensor.
        (b) Reference distance, sensor distance, and estimation error.}
    \label{fig:sensor_perf}
\end{figure}
\begin{figure}[t]
    \centering
    \includegraphics[width=\columnwidth]{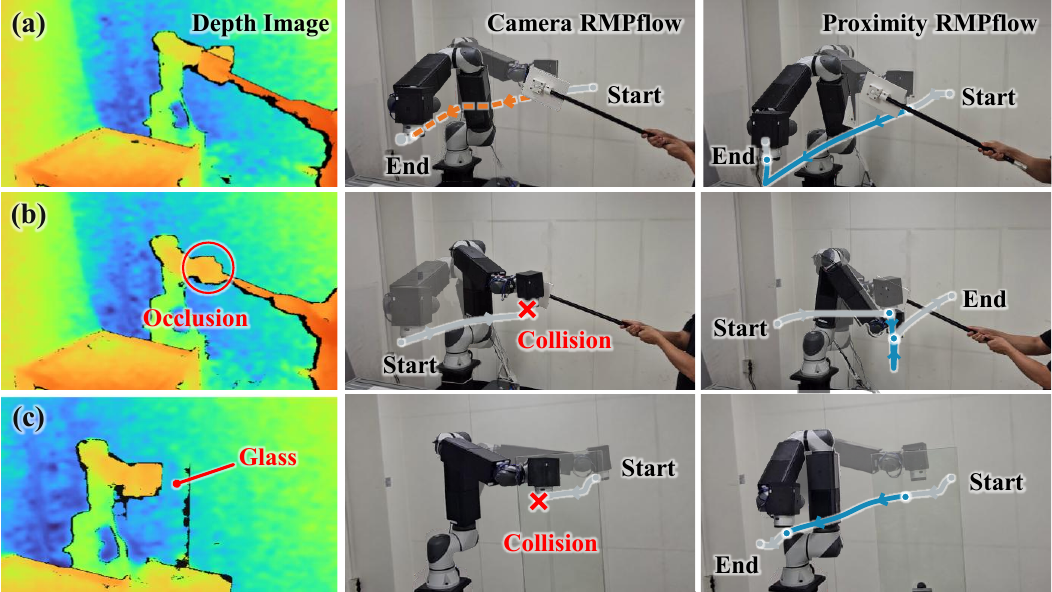}
    \caption{RMPflow avoidance using camera and proximity sensing under
        (a) a suddenly appearing obstacle (SAO), (b) SAO with
        robot-induced occlusion, and (c) transparent glass.
        The leftmost column shows depth images.
        The solid grey line denotes the \termfix{base RMPflow trajectory},
        the dashed orange line the \termfix{camera-based RMPflow path},
        and the solid blue line the \termfix{proximity-based RMPflow path}.}
    \label{fig:sensor_compare}
\end{figure}

\begin{figure*}[t]
    \centering
    \color{black}
    \includegraphics[width=\textwidth]{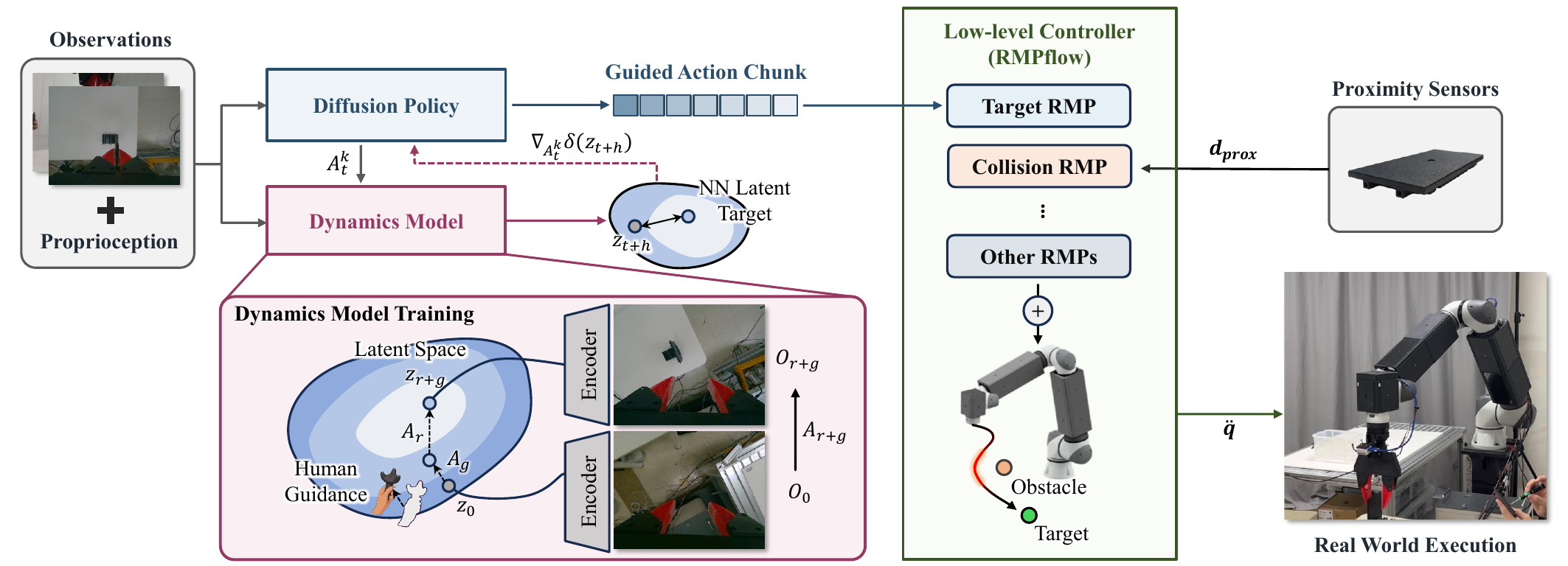}
    \caption{Integrated IDP architecture in which Diffusion Policy
        generates reference targets, RMPflow executes them safely, and
        LPB guides recovery from avoidance-induced OOD states.
        \rev{The inset shows how the dynamics model is trained. An
        action sequence $A_{r+g}$ that combines the policy's own
        rollout actions $A_r$ with the human guidance
        actions $A_g$ carries the latent state from $z_0$ to
        $z_{r+g}$, so the training transitions reach beyond the expert
        region of the latent space and cover the off-path states that
        RMPflow avoidance produces.}}
    \label{fig:overview}
\end{figure*}

\subsection{Self-Detection Problem}
Proximity sensors mounted on robot links may detect other links and
mounting structures in addition to external objects. Changes in joint
configuration alter the relative positions of these structures and the
sensors, so the robot's own response may be mistaken for an approaching
external object. To address this, the method of Moon
et al.~\cite{moon2021dual} is applied to compensate for self-detection.
After the sensors are mounted, responses are collected at various joint
configurations to learn the contribution from the robot itself. During
execution, sensor measurements are corrected using the learned response
for the current configuration. The corrected distance estimates are
used as obstacle input to RMPflow.
\endgroup
\begingroup
\color{black}
\section{Interactive Diffusion Policy}
\label{sec:method}

\rev{%
This section describes how IDP integrates task execution,
proximity-based avoidance, and recovery from avoidance-induced OOD
states. It then presents the data collection procedure used to train
the LPB dynamics model.}

\subsection{Imitation Learning with Diffusion Policy}
\label{sec:dp}
A visuomotor policy is trained on expert demonstrations to predict a
reference end-effector target conditioned on the current observation
$\mathbf{o}_t$, which consists of RGB images from one wrist-mounted
camera and one external camera and the proprioceptive state.
\rev{We use Diffusion Policy~\cite{chi2023diffusion},
which represents the conditional action distribution through iterative
denoising.} Training perturbs an expert action sequence
$\mathbf{A}_t^{0}$ with noise $\boldsymbol{\epsilon}^{k}$ at a random
diffusion step $k$ and learns a noise-prediction network
$\epsilon_\theta$ by minimizing
\begin{equation}
    \color{black}
    \mathcal{L} =
    \mathbb{E}_{k,\,\boldsymbol{\epsilon}^{k}}
    \big\| \boldsymbol{\epsilon}^{k} -
    \epsilon_\theta\big(\mathbf{o}_t,\,
    \mathbf{A}_t^{0} + \boldsymbol{\epsilon}^{k},\, k\big)
    \big\|_2^{2}
    \label{eq:dp_loss}
\end{equation}
At run time, an action sequence is generated from Gaussian noise
$\mathbf{A}_t^{K}$ by $K$ denoising iterations,
\begin{equation}
    \mathbf{A}_t^{k-1} = \alpha_k\big(\mathbf{A}_t^{k}
        - \gamma_k\,\epsilon_\theta(\mathbf{o}_t,
          \mathbf{A}_t^{k}, k)\big)
        + \mathcal{N}\big(0, \sigma_k^{2}\mathbf{I}\big)
    \label{eq:dp_denoise}
\end{equation}
where $\alpha_k$, $\gamma_k$, and $\sigma_k$ follow the noise
schedule. The first portion of the denoised sequence is executed as
the reference target $\mathbf{x}_t^{\text{ref}}$ before replanning.

\endgroup

\begingroup
\color{black}
\subsection{RMPflow-Based Control}
\label{sec:rmp}
RMPflow~\cite{cheng2020rmpflow} is a framework that combines motion policies defined in different task spaces using state-dependent weighting matrices to incorporate multiple motion objectives into a single robot motion. Each policy, represented by its desired acceleration \(\mathbf a_i\) and weighting matrix \(\mathbf M_i\), is mapped to joint space to compute the resulting joint acceleration as follows.

\begin{equation}
\label{eq:rmp}
\ddot{\mathbf q}
=
\left(
\sum_i \mathbf J_i^{\mathsf T}\mathbf M_i\mathbf J_i
\right)^{\dagger}
\sum_i
\mathbf J_i^{\mathsf T}\mathbf M_i
\left(
\mathbf a_i-\dot{\mathbf J}_i\dot{\mathbf q}
\right)
\end{equation}

Here, \(\mathbf q\) denotes the joint positions, \(\mathbf J_i\) is the Jacobian mapping joint space to the task space of policy \(i\), and \(\dot{\mathbf J}_i\dot{\mathbf q}\) is the correction term for acceleration coordinate transformations. The symbol \(\dagger\) denotes the pseudoinverse. This combination can be interpreted as minimizing the error between the acceleration requested by each policy and the actual task space acceleration, weighted by \(\mathbf M_i\).

In this study, RMPflow is used as a low-level controller to integrate tracking of targets generated by Diffusion Policy with obstacle avoidance based on proximity sensing.
The Target RMP receives the end-effector target generated by Diffusion Policy and produces a tracking acceleration that reduces the error between the current state and the target. The Collision RMP incorporates distance measurements from the proximity sensors as obstacle information and produces an avoidance acceleration to maintain separation from detected obstacles. As an obstacle approaches, the contribution of the avoidance policy increases and adjusts the tracking motion. Thus, task execution and avoidance are handled within the same control structure without switching between the two policies.
\endgroup

\begin{figure}[t]
    \centering
    \color{black}
    \includegraphics[width=0.9\columnwidth]{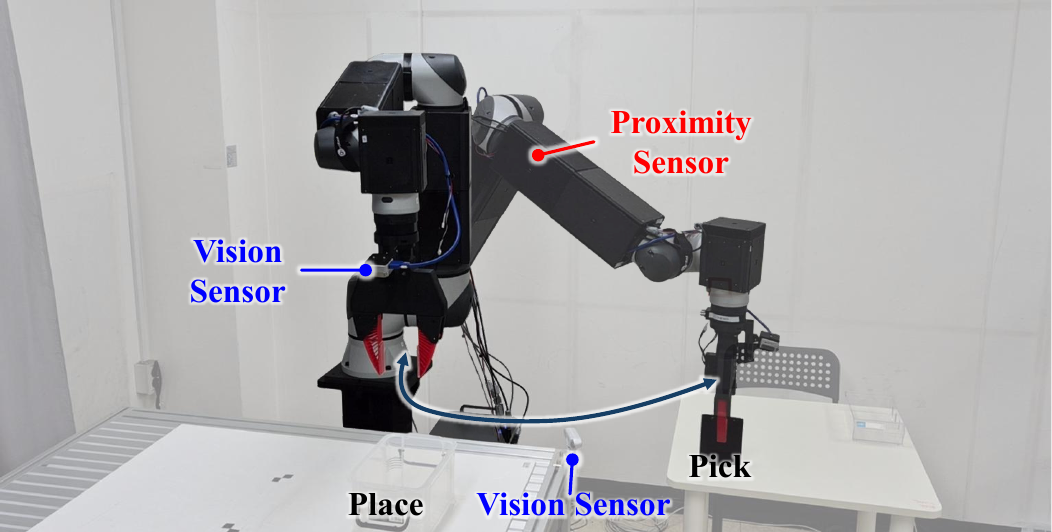}
    \caption{Experimental setup for pick-and-place showing the RB10
        robot, vision cameras, and proximity sensors.}
    \label{fig:experimental_setup}
\end{figure}

\begingroup
\color{black}
\subsection{Handling Avoidance-Induced OOD with LPB}
\label{sec:lpb}
Avoidance maneuvers move the robot to configurations the policy
never observed during training, so naive resumption lets imitation
errors compound. We adopt the Latent Policy Barrier
(LPB)~\cite{lee2025latentpolicybarrier}, which treats the latent
embeddings of the expert demonstrations as an implicit barrier
between in-distribution states the policy can handle and
OOD states where it is unreliable. At every
step we encode the current observation as $\mathbf{z}_t =
E(\mathbf{o}_t)$\rev{, where $E$ denotes the frozen visual encoder of
Diffusion Policy}. Let
$\mathcal{Z}_{\mathrm{exp}}$ denote the set of latent embeddings of
expert observations. The \termfix{OOD cost} is defined as
the squared distance to the nearest expert embedding,
\begin{equation}
    \color{black}
    \begin{split}
        \mathcal{Z}_{\mathrm{exp}}
        &= \{E(\mathbf{o}_i) \mid
             \mathbf{o}_i \in \mathcal{D}_{\mathrm{exp}}\} \\
        \delta(\mathbf{z}_t)
        &= \min_{\mathbf{z}_i \in \mathcal{Z}_{\mathrm{exp}}}
           \|\mathbf{z}_t - \mathbf{z}_i\|_2^2
    \end{split}
    \label{eq:ood_score}
\end{equation}
When $\delta(\mathbf{z}_t)$ exceeds a threshold $\tau$, LPB steers the diffusion denoising through its
learned dynamics model $d_\phi$ toward actions whose predicted
future latent re-enters the expert manifold,
\begin{equation}
    \hat{\epsilon}(\mathbf{A}_t^k) = \epsilon_\theta(\mathbf{A}_t^k)
        - \eta\sqrt{1-\bar{\alpha}_k}\,
          \nabla_{\mathbf{A}_t^k}\,
          \delta\big(d_\phi(\mathbf{z}_t, \mathbf{A}_t^k)\big)
    \label{eq:lpb}
\end{equation}
where $\mathbf{A}_t^k$ is the action sequence sample at denoising iteration
$k$, and $\eta$ is the guidance scale. We follow the dynamics-model
training formulation of LPB~\cite{lee2025latentpolicybarrier}, using
recovery data that include RMPflow avoidance and human-guided task
resumption, as described in Section~\ref{sec:datacollect}.

\subsection{Framework}
\label{sec:framework}
Fig.~\ref{fig:overview} illustrates the overall IDP framework for executing a learned manipulation task in a workspace shared with a human collaborator.
A single-arm manipulator with end-effector
pose $\mathbf{x}_t\in\mathrm{SE}(3)$ executes a visuomotor policy
$\pi_\theta(\mathbf{x}^{\text{ref}}_t \mid \mathbf{o}_t)$ trained on
expert demonstrations. A dynamic human collaborator shares the
workspace. The safety layer observes the collaborator through
robot-mounted proximity readings $\mathbf{s}_t\in\mathbb{R}^{N_p}$
(Section~\ref{sec:proximity}).
\rev{The objective is to carry out the behavior learned from the
expert demonstrations while maintaining a safe distance from the
collaborator throughout execution.}

\rev{Diffusion Policy predicts reference targets from visual and
proprioceptive observations. The RMPflow controller runs continuously
underneath, tracking the reference when the workspace is clear and
deflecting the arm when the proximity channels detect an approaching
obstacle. LPB runs in parallel, monitoring the latent \termfix{OOD cost} and
engaging its guidance once the \termfix{cost} exceeds the threshold $\tau$.}
Algorithm~\ref{alg:control} summarizes the deployment loop, where $\ddot{\mathbf{q}}_t$ denotes the
joint acceleration command.
\rev{The RMPflow controller is evaluated at every step and is the only
consumer of the proximity signal, while LPB guidance is gated by the
latent \termfix{OOD cost} alone, so avoidance and recovery are triggered
independently. Fig.~\ref{fig:overview} also illustrates how the dynamics model is
trained on rollouts that include RMPflow avoidance and human-guided task
resumption. These data cover off-path states induced by avoidance and
the transitions back toward task execution.}

\begin{algorithm}[t]
\color{black}
\caption{\methodabbr{} control loop}\label{alg:control}
\begin{algorithmic}[1]
\While{task not done}
    \State $\mathbf{o}_t \gets \texttt{observe}()$
    \State $\mathbf{s}_t \gets \texttt{proximity\_sense}()$
    \State $\mathbf{z}_t \gets E(\mathbf{o}_t)$
           \Comment{\rev{frozen policy encoder}}
    \State $\delta_t \gets \termfix{\texttt{ood\_cost}}(\mathbf{z}_t,
           \mathcal{Z}_{\mathrm{exp}})$
           \Comment{squared nearest-neighbor distance}
    \If{$\delta_t > \tau$} \Comment{OOD detected}
        \State $\mathbf{x}_t^{\text{ref}} \gets
               \pi_\theta(\mathbf{o}_t)$ with LPB
               guidance (Eq.~\eqref{eq:lpb})
    \Else
        \State $\mathbf{x}_t^{\text{ref}} \gets
               \pi_\theta(\mathbf{o}_t)$
    \EndIf
    \State $\ddot{\mathbf{q}}_t \gets
           \pi_{\mathrm{RMPflow}}(\mathbf{x}_t^{\text{ref}}, \mathbf{s}_t)$
           \Comment{proximity-driven avoidance}
    \State $\texttt{execute}(\ddot{\mathbf{q}}_t)$
\EndWhile
\end{algorithmic}
\end{algorithm}

\subsection{Data Collection}
\label{sec:datacollect}
Data are collected on the RB10 equipped with RGB cameras and proximity
sensors, using the pick-and-place setup shown in Fig.~\ref{fig:experimental_setup}.
The collection procedure produces base expert demonstrations
$\mathcal{D}_{\mathrm{exp}}$ and recovery trajectories
$\mathcal{D}_{\mathrm{rec}}$. The base demonstrations train the
visuomotor policy and define its expert latent set. The dynamics model
for LPB is trained with fixed-horizon slices of both the base
demonstrations and the recovery trajectories. The resulting dataset
and dynamics-model objective are
\begin{equation}
    \color{black}
    \begin{aligned}
        \mathcal{D}_{\mathrm{dyn}}
        &= \operatorname{Slice}_{h}(\mathcal{D}_{\mathrm{exp}})
           \cup \operatorname{Slice}_{h}(\mathcal{D}_{\mathrm{rec}}) \\
        \mathcal{L}_{\mathrm{dyn}}(\phi)
        &= \mathbb{E}_{\xi_t \sim \mathcal{D}_{\mathrm{dyn}}}
           \left[\left\|
           E(\mathbf{o}_{t+h})
           - d_\phi\big(E(\mathbf{o}_t),\mathbf{A}_t\big)
           \right\|_2^2\right]
    \end{aligned}
    \label{eq:dyn_loss}
\end{equation}
where \rev{$\operatorname{Slice}_{h}$ cuts each trajectory into
samples $\xi_t = (\mathbf{o}_t,\mathbf{A}_t,\mathbf{o}_{t+h})$, in
which the action sequence $\mathbf{A}_t$ separates the two
observations by the prediction horizon $h$}.

\begin{figure}[t]
    \centering
    \color{black}
    \includegraphics[width=\columnwidth]{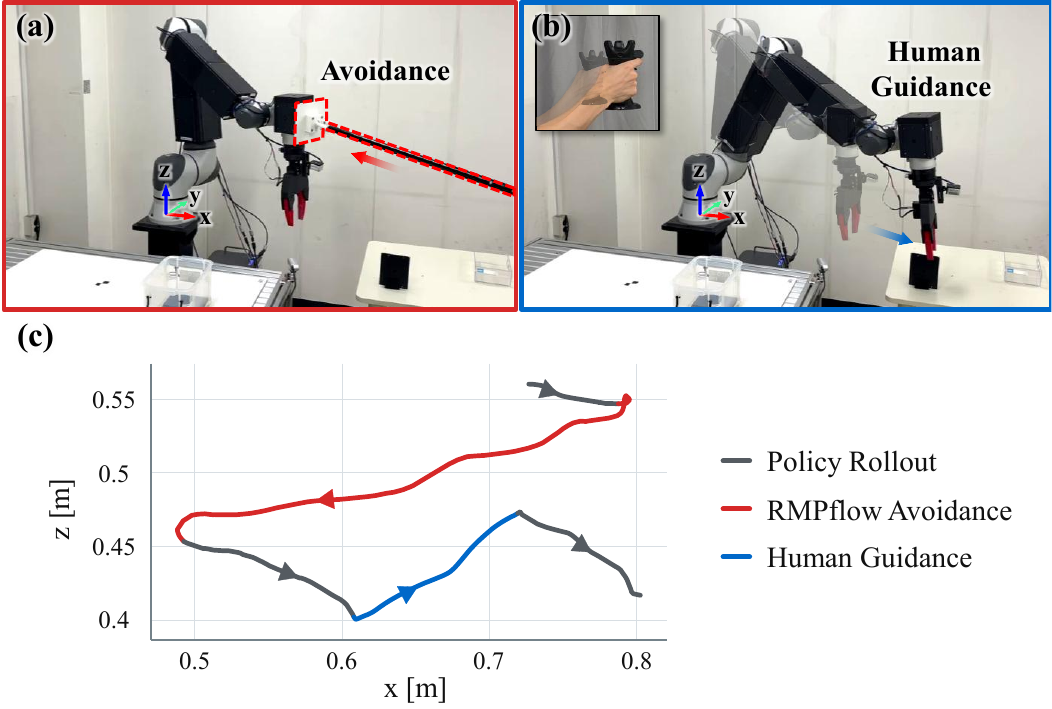}
    \caption{
        \rev{Recovery-aware data collection. A checkpoint rollout is
        deflected by RMPflow when an obstacle approaches, and the operator
        provides human guidance when the policy cannot resume from
        the resulting off-path state. The rollout is recorded for
        training the LPB dynamics model.}}
    \label{fig:datacollect}
\end{figure}

The base dataset contains
successful task demonstrations recorded under normal conditions with
no obstacle entering the workspace. These demonstrations train
Diffusion Policy and provide the expert latent set used by LPB.

\rev{Recovery trajectories are
collected while the policy itself drives the robot. Rollouts from intermediate policy checkpoints can reach states outside the expert demonstration distribution even without obstacle-induced avoidance. The disturbance, however, does not come from
the checkpoint itself. An obstacle approaches the robot at
unpredictable moments, and the RMPflow controller deflects the arm
along the avoidance motion computed from the proximity readings. Once
the obstacle has left, the policy resumes the task on its own whenever
the deflection has left it close enough to the expert distribution.
When the deflection has instead moved it into an OOD state and the
task stalls, the operator provides human guidance
until execution continues.} Fig.~\ref{fig:datacollect} visualizes
the RMPflow deflection,
\rev{human guidance}, and matched end-effector trajectories.

Delegating the avoidance to RMPflow rather than to the human
teleoperator brings several practical benefits. The operator is
relieved of reacting to the intruding obstacle and can concentrate on
the task itself, which makes long data collection sessions less
fatiguing and less error-prone. Because the avoidance is produced by
a fixed controller, the deflection motions are consistent and
repeatable across episodes instead of varying with each operator's
reflexes, so the recovery data reflect the same avoidance dynamics
that the policy will face at deployment.

\endgroup
\section{Experiments}
\label{sec:exp}
\rev{The experiments evaluate whether the robot avoids contact with intruding obstacles and resumes and completes the learned task after avoidance.}

\subsection{Experimental Setup}
We use a single-arm Rainbow Robotics
\rev{RB10 manipulator} with a parallel-jaw gripper. \rev{RGB cameras
provide the policy's observation.} The proximity sensors are installed
as described in Section~\ref{sec:proximity}.
Fig.~\ref{fig:experimental_setup} shows the complete experimental
setup.

The robot performs a
\rev{wide-range pick-and-place task while obstacles intrude into its
workspace at random moments. The disturbances take two forms.
A dynamic global blocking (DGB) obstacle moves continuously and drives
the robot away from its nominal trajectory, whereas a suddenly
appearing obstacle (SAO) requires the robot to avoid it quickly and
then return to the task.}

\rev{The base dataset consists of 100 teleoperated demonstrations
with no obstacle entering the workspace. The recovery dataset was
collected on the same task from rollouts of intermediate policy
checkpoints under obstacle intrusion, following the procedure of
Section~\ref{sec:datacollect}. Each rollout was logged in full, so a
recovery trajectory contains the RMPflow avoidance motion together
with the resumption that follows it, whether the policy resumed on
its own or under human guidance.}

\begin{figure}[t]
    \centering
    \color{black}
    \includegraphics[width=\columnwidth]{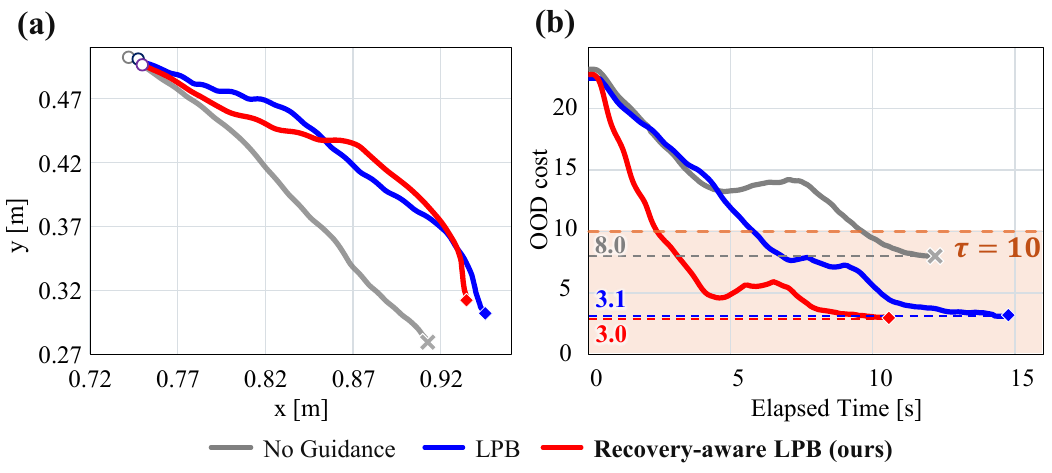}
    \caption{Recovery from matched off-path states. (a) End-effector
        trajectories and (b) \termfix{OOD cost} for No Guidance, LPB with the
        dynamics model trained on checkpoint rollouts, and
        Recovery-aware LPB, whose dynamics model is trained on
        data collected under actual RMPflow avoidance. The orange-shaded
        region indicates OOD costs below the LPB activation threshold
        ($\tau=10$).}
    \label{fig:lpb_vis}
\end{figure}

\begin{table}[!t]
\centering
\color{black}
\caption{\rev{Task completion success from post-avoidance states for the three recovery conditions}}
\label{tab:dataset}
\renewcommand{\arraystretch}{1.35}
\setlength{\tabcolsep}{9pt}
\begin{tabular}{lc}
\toprule
\rev{Condition} & Success rate $\uparrow$ \\
\midrule
\rev{No Guidance} & \rev{25.0\%} \\
\rev{LPB} & \rev{63.8\%} \\
\rev{\textbf{Recovery-aware LPB (ours)}} & \rev{\textbf{91.3\%}} \\
\bottomrule
\end{tabular}
\end{table}

\begin{figure*}[t]
    \centering
    \includegraphics[width=\textwidth]{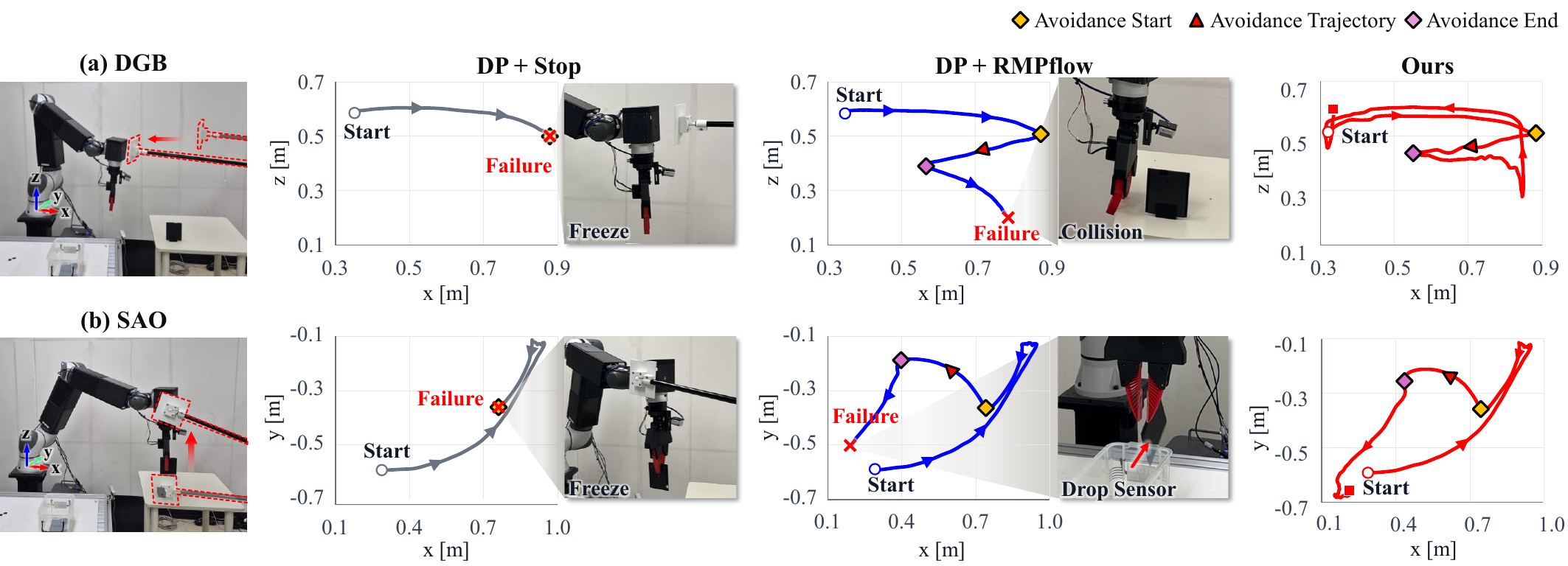}
    \caption{Comparison of end-effector trajectories during obstacle
        avoidance and task resumption under (a) dynamic global blocking
        (DGB) and (b) suddenly appearing obstacles (SAO). The black
        trajectory shows DP$+$Stop, the blue trajectory shows
        \termfix{RMPflow-based avoidance} without LPB, and the red trajectory
        shows our method with Recovery-aware LPB.}
    \label{fig:method_compare}
\end{figure*}

\begin{figure*}[t]
    \centering
    \color{black}
    \includegraphics[width=\textwidth]{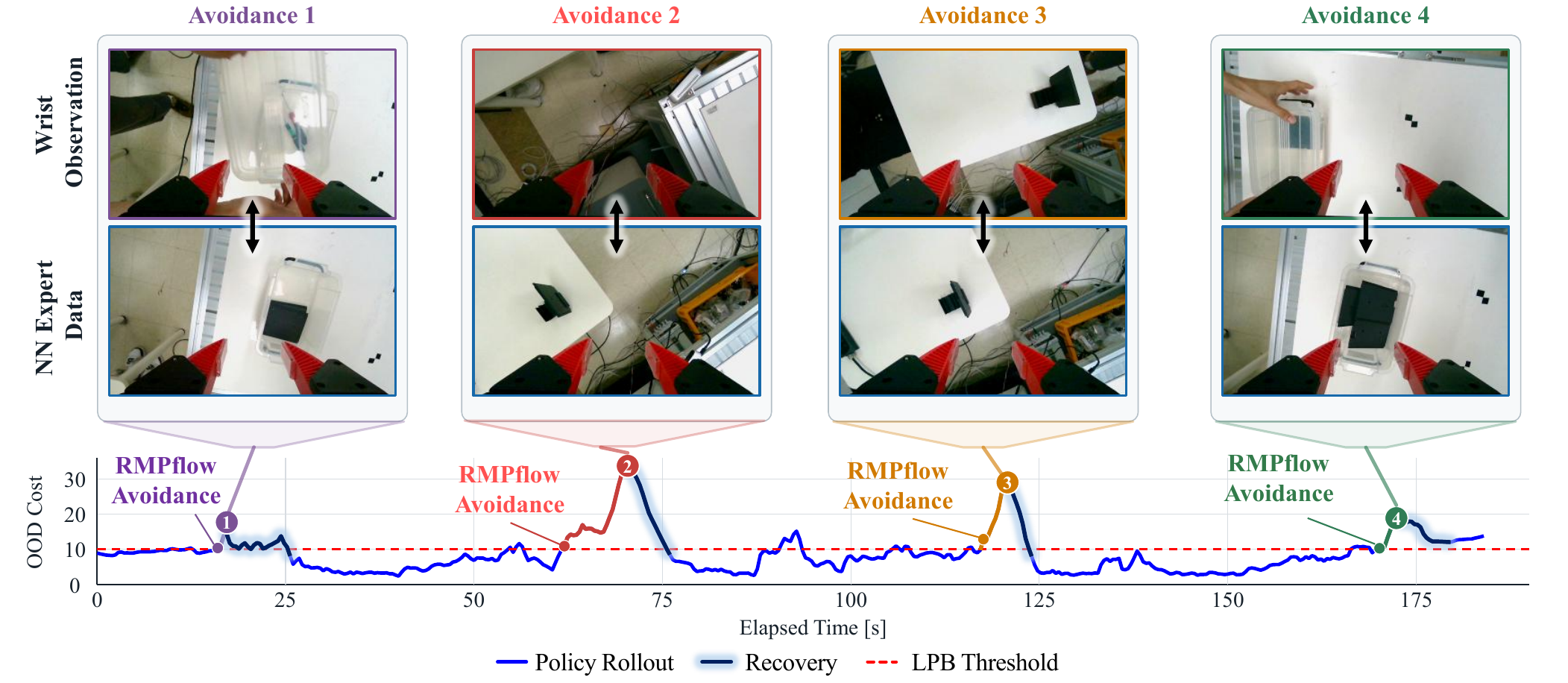}
    \caption{Avoidance-induced OOD events during a representative
        pick-and-place rollout. Each event pairs the RMPflow response
        with the current wrist observation and its nearest expert image.
        The lower plot shows the latent \termfix{OOD cost} relative to
        the activation threshold.}
    \label{fig:rollout}
\end{figure*}

\subsection{Recovery Dataset Ablation}
\label{sec:ablation_dataset}
\rev{This study tests} whether data collected during avoidance and
subsequent resumption improve the LPB dynamics model training.
\rev{It uses the same pick-and-place task, and each trial starts from a post-avoidance state. The full \methodabbr{} architecture is held fixed and only
the recovery mechanism varies, as listed in Table~\ref{tab:dataset}.}
\begin{enumerate}
    \item \rev{\textbf{No Guidance} disables LPB, so the policy
          resumes from the deflected state on its own.}
    \item \rev{\textbf{LPB}} uses \rev{a dynamics model trained}
          on rollouts from intermediate policy checkpoints, following
          LPB~\cite{lee2025latentpolicybarrier}.
    \item \rev{\textbf{Recovery-aware LPB (ours)}} \rev{uses a
          dynamics model trained on the recovery trajectories of
          Section~\ref{sec:datacollect}.}
\end{enumerate}
\rev{Each trial is scored on whether the robot completes the task from a
post-avoidance state, so the comparison measures task resumption rather
than nominal execution. Each condition is evaluated from eight robot poses representing post-avoidance states, with 10 trials per pose. Without guidance the policy completes the task
in $25.0\%$ of the trials. Training the dynamics model on checkpoint
rollouts raises the success rate to $63.8\%$, and training it on the
recovery trajectories reaches $91.3\%$. The gap between the last two
conditions suggests that the dynamics model benefits from data
covering the states that RMPflow avoidance produces, which checkpoint
rollouts alone do not reach.} Fig.~\ref{fig:lpb_vis} complements the
task-level comparison with end-effector trajectories and latent \termfix{OOD cost} for
the three \rev{conditions. All three runs start from the same
deflected state. With the recovery data the \termfix{OOD cost} falls below the
activation threshold earliest and remains below it for the rest of the
episode, whereas the rollout data require a substantially longer
interval to reach a comparable level. Without guidance the cost eventually falls below $\tau$ but
settles at 8.0, well above the level reached by the two guided runs, and the
corresponding trajectory in Fig.~\ref{fig:lpb_vis}(a) departs from the demonstrated path and
does not complete the task}.

\subsection{Recovery Under Dynamic Obstacles}
\label{sec:main_exp}
\rev{This study examines how the robot behaves under the DGB and SAO
disturbances. The three conditions share the same proximity sensing
and differ in how the robot reacts once an obstacle is detected, as
compared in Fig.~\ref{fig:method_compare}. In Fig.~\ref{fig:method_compare}, yellow diamonds
mark the initiation of avoidance when an obstacle appears, red arrows indicate
the avoidance trajectories, and purple diamonds mark the end of avoidance.}
\begin{enumerate}
    \item \rev{\textbf{DP$+$Stop} halts the arm while the obstacle
          stays within the safety distance and continues the task
          once the workspace is clear.}
    \item \rev{\textbf{DP$+$\termfix{RMPflow}} deflects the arm with the
          RMPflow controller and leaves the policy to continue on its
          own once the obstacle has passed.}
    \item \rev{\textbf{DP$+$\termfix{RMPflow}$+$Recovery-aware LPB (\methodabbr{})} engages
          LPB guidance while the \termfix{OOD cost} stays above the threshold,
          so the policy is steered back toward the expert manifold
          before it resumes the task. It uses the dynamics model
          trained on the recovery trajectories in
          Section~\ref{sec:datacollect}.}
\end{enumerate}
\rev{In the representative trials of Fig.~\ref{fig:method_compare}, both baselines fail under either disturbance. DP$+$Stop keeps the arm clear of the obstacle, but the obstacle remains
within the proximity range that triggers a stop, so the arm remains
stationary until the episode times out. DP$+$\termfix{RMPflow} avoids the
intruding obstacle in both scenarios, yet the deflected state leaves
the policy in an OOD state, and continuing from there the arm
strikes the workspace under DGB and drops the grasped object under
SAO. \methodabbr{} instead avoids the obstacle without contact and completes
the task after the avoidance in both scenarios.}

\rev{Fig.~\ref{fig:rollout} follows the latent \termfix{OOD cost} through a
single \methodabbr{} rollout. Each avoidance pushes the arm off the
demonstrated path and the cost climbs until it crosses the activation
threshold, at which point the guidance engages and drives the cost
back down. The wrist observation then matches the nearest expert image
again and the rollout continues at a low cost until the next
intrusion, and the same cycle repeats at every avoidance event.}

\begingroup
\color{black}
\section{Conclusion}
In this paper, Interactive Diffusion Policy (IDP) is proposed so that a robot sharing its workspace with people keeps a safe distance from an approaching obstacle and still carries the learned task through to completion.
IDP lets Diffusion Policy generate reference targets and an RMPflow controller track them while robot-mounted proximity sensing keeps the arm away from obstacles.
Because the deflection leaves the policy in an OOD state, the Latent Policy Barrier (LPB) guides it back toward the expert manifold, using a dynamics model trained on recovery data produced by actual avoidance.
In the representative trials under both disturbances, IDP avoids the obstacle and completes the task, whereas both baselines fail.
In the evaluation from post-avoidance states, task completion success increases from $25.0\%$ without guidance to $91.3\%$ with Recovery-aware LPB.
The dataset ablation shows that adding recovery data collected under actual avoidance raises the resumption success rate from $63.8\%$ to $91.3\%$.
Future work will evaluate IDP with multiple people in the workspace and across a wider range of tasks.
\endgroup
\bibliographystyle{IEEEtran}
\bibliography{references}

\end{document}